# Simplified Swarm Optimisation for the Hyperparameters of a Convolutional Neural Network

Wei-Chang Yeh[1], Yi-Ping Lin[1], Yun-Chia Liang[2], Chyh-Ming Lai[3], and Xiao-Zhi Gao[4]
[1] Department of Industrial Engineering and Engineering Management, National Tsing Hua University, Taiwan.
[2] Industrial Engineering and Management, Yuan Ze University, Taiwan.
[3] Management College, National Defense University, Taiwan.
[4] School of Computing, University of Eastern Finland, Kuopio, Finland

Convolutional neural networks (CNNs) are widely used in image recognition. Numerous CNN models, such as LeNet, AlexNet, VGG, ResNet, and GoogLeNet, have been proposed by increasing the number of layers, to improve the performance of CNNs. However, performance deteriorates beyond a certain number of layers. Hence, hyperparameter optimisation is a more efficient way to improve CNNs. To validate this concept, a new algorithm based on simplified swarm optimisation is proposed to optimise the hyperparameters of the simplest CNN model, which is LeNet. The results of experiments conducted on the MNIST, Fashion MNIST, and Cifar10 datasets showed that the accuracy of the proposed algorithm is higher than the original LeNet model and PSO-LeNet and that it has a high potential to be extended to more complicated models, such as AlexNet.

Corresponding author: W.C. Yeh; email: yeh@ieee.org

## 1. INTRODUCTION

Deep learning, a special machine learning method based on an artificial neural network (ANN), has recently gained popularity. Deep learning makes a machine intelligent such that a computer can simulate the functions of the human brain to observe, analyse, learn human behaviour, and make decisions.

Among deep learning applications, including image recognition [1], artistic creation [2], semantic understanding [3], and poetry creation [4], image recognition has become the most popular research field in recent years. In addition, image recognition is an important task that can be applied to transportation, home, manufacturing, and medical applications, such as autonomous driving [7], healthcare [8], product defect detection [9], and medical imaging [10, 11], making people's lives more convenient.

Convolutional neural networks (CNNs) are the most intensively researched [12] models for image recognition because they are more accurate than human judgement [13]. With the combination of three layers (that is, the convolution, pooling, and fully connected layers) and one function, CNNs have considerable flexibility to allow users to make modifications according to their needs.

The history of CNNs can be traced back to 1962 [14]; however, the model that is closest to the present definition of a CNN is the LeNet proposed by Yann LeCun in 1989, and it has been revised repeatedly since then [15]. After the proposal of AlexNet [1] in 2012, there have been significant advancements in CNNs. Many CNN models, such as VGG, ResNet, and GoogLeNet, have been developed successively [17-20].

Increasing the number of layers leads to less accurate results [21]. Hence, numerous studies have investigated methods to improve CNN performance without changing the architecture. Hyperparameters include the size of kernels, number of kernels, length of strides, and pooling size, which directly affect the performance and training speed of CNNs. Moreover, the impact of hyperparameters increases as the complexity of the network increases. Hence, the most popular method of improving CNN performance is to optimise the hyperparameters [22, 23].

CNN hyperparameter optimisation is an integer programming problem. In the past, many studies used manual designs for hyperparameters; that is, scholars or experts must adjust the hyperparameters based on experience and expertise, which is not only unfounded but also time consuming during testing. Several heuristic approaches, such as a grid search, randomised search [27], Bayesian optimisation, and gradient-based optimisation, have been developed for hyperparameter optimisation. The drawback of the above methods is that they are less efficient in a high-dimensional space because the number of evaluations increases exponentially as the number of hyperparameters increases [28-30].

Hence, artificial intelligence techniques, including the genetic algorithm [34-37], particle swarm algorithm (PSO) [38-40], and artificial bee colony algorithm [41], have been proposed for this problem. However, most of these methods change the CNN structure or combine different algorithms to optimise the performance and are complicated and difficult for users to understand. Thus, there is a need for a simple algorithm that does not change the CNN structure for the hyperparameters.

The purpose of this study is to apply the simplified swarm optimisation (SSO) algorithm proposed by Yeh to tune the CNN hyperparameters [6]. The SSO is not only simple and easy to understand but also efficient. Many studies have demonstrated the excellent ability of SSO in optimisation problems [42-46]; however, no research has applied SSOs to the hyperparameter optimisation problem. Therefore, in this study, a new algorithm called the SSO-LeNet is proposed to apply SSO to the original LeNet architecture without changing the layers and validate it with different datasets for automated hyperparameter optimisation.

The remainder of this paper is organised as follows. Section 2 provides an overview of CNNs, LeNet, and SSO, which are the basis of the proposed algorithm. Section 3 describes the major parts of the proposed SSO-LeNet, including the special solution structure, fitness function, sequential dynamic variable range (SDVR), and small-sample design matrix. Section 4 provides the pseudocode and flowchart for further details. Section 5 describes three experiments: Ex1, Ex2, and Ex3, on three benchmark datasets, including NMIST, Fashion-MNIST, and CIFAR10, to demonstrate ways to use the proposed SDVR and the small-sample design matrix in Ex1, and then compares the proposed SSO-LeNet to the traditional LeNet [49] and PSO-LeNet [39] in Ex2 and Ex3, respectively. Finally, Section 6 concludes the study and provides possible future work.

## 2. OVERVIEW OF ANN, CNN, LENET, AND SSO

ANNs are the basis of CNNs and LeNet [49] is the simplest CNN that we propose an SSO-based algorithm to optimise its hyperparameters. Hence, ANNs, CNNs, LeNets, and SSOs are reviewed in this section before introducing the proposed SSO-LeNet.

### 2.1 Artificial Neural Network

An ANN, also known as a multi-layer perceptron, is a special network model comprising many nodes and arcs, as shown in Fig. 1. These nodes mimic human nerve cells, called neurons, and transmit information to the next layer by connecting neurons in different layers. Every neuron in each layer is connected to all the neurons in the next layer; however, neurons in the same layer are not connected.

As the depth of a neural network increases, a deeper ANN with more hidden layers, called a deep neural network, can be used to solve problems that are more complex [5, 6].

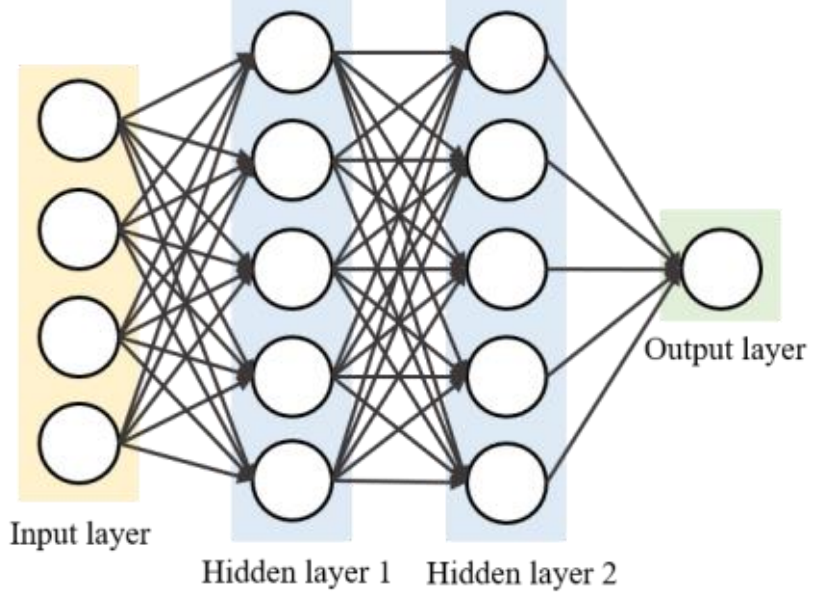

**Figure 1.** ANN Structure

### 2.2 Convolutional Neural Network

A CNN is developed from an ANN, as shown in Fig. 2; therefore, it also has an input layer, hidden layers, and an output layer. To extract features and classify images, four main operations are used to build a CNN model. These operations are described in detail in Sections 2.2.1–2.2.4.

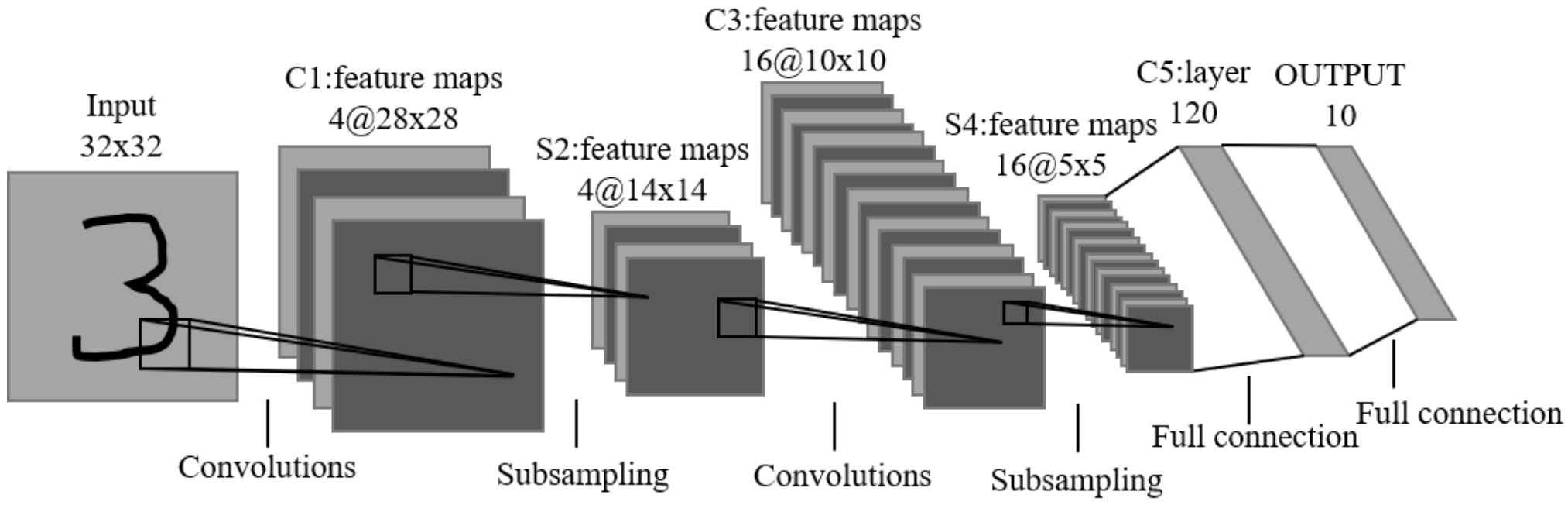

**Figure 2.** LeNet-4 [50].

### 2.2.1 Convolution Layer

The convolution layer, as shown in the first and third layers in Fig. 2, is used for image feature extraction through multiple convolution kernels (filters), which can mine the abstract information in the original data, such as edge detection, blurring, sharpening, or embossing.

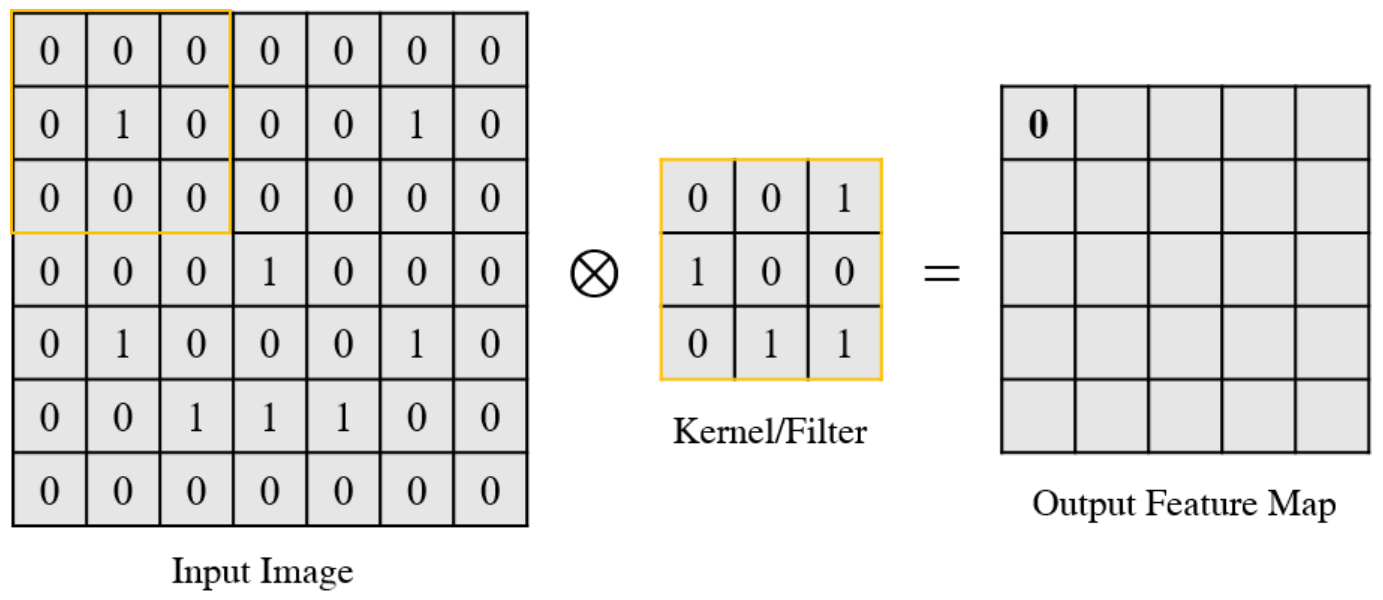

**Figure 3.** Convolution process

The convolution process is shown in Fig. 3, in which the sum of an element-by-element product is applied to the weights of both the convolution kernel (refer to 'Kernel/Filter' in Fig. 3) and input map (refer to 'Input Image' in Fig. 3). Then slides on the feature map according to the strides and continues to convolve until the entire feature map is convolved to obtain the weight of the output feature map.

The relationship between the input image size and output image size can be derived from Eq. (1).

$$O = floor(\frac{I - K + 2P}{S}) + 1 \tag{1}$$

Here,

- $O$ and $I$ denote the size of the output and input images, respectively;
- $K$ represents the size of the kernel;
- $P$ indicates the number of zero-padding that fills the boundary of the feature image with zero weight; and
- $S$ is the symbol for the size of the stride, which is the number of pixel shifts over the input image.

For example, in Fig. 2, let the size of the input image be $5 \times 5$, the size of the convolution kernel be $3 \times 3$, the stride be 1, and the zero-padding be used. The size of the output feature after convolution is $5 \times 5$.

### 2.2.2 Pooling Layer

Generally, the convolution layer outputs a feature map to the pooling layers, as depicted in the second and fourth layers of Fig. 2, which is also known as the subsampling layer. Pooling layers avoid overfitting by pooling the convolved feature map, decreasing the dimensionality of the feature map, reducing sampling, and retaining important features.

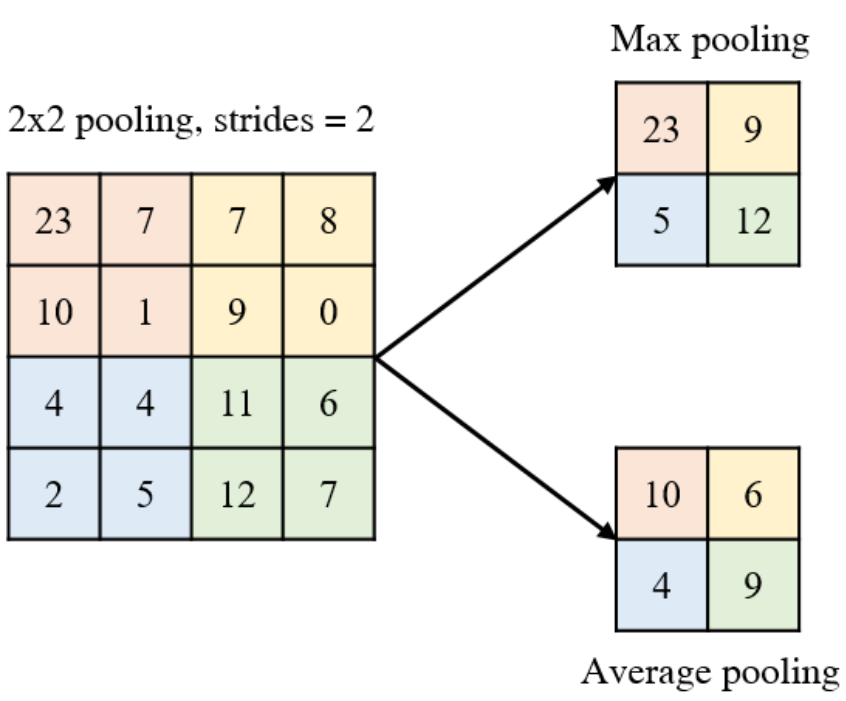

**Figure 4.** Pooling process

There are two common pooling methods: average pooling and maximum pooling, which operate similarly to convolution. The size and strides of the pooling window must first be determined. The pooling process is shown in Fig. 4.

Pooling considers the average or maximum value of a local block; slight distortions in the input image do not affect the output image, and the output can be obtained in almost the same proportion as the input image.

In addition, the output size of the feature map after the pooling layer can be obtained using Eq. (1).

### 2.2.3 Fully Connected Layer

As demonstrated in Fig. 2, the first half of the CNN comprises multiple convolution layers (i.e., the first and the third layers) and pooling layers (i.e., the second and fourth layers) alternately for extracting and learning features. The second half is created after flattening (which is the input layer of the fully connected layer; refer to the first 'Full connection' in Fig. 2).

The second half comprises a fully connected layer and an output layer for image classification. The fully connected layer is similar to the traditional ANN in which the neurons in each layer are connected to all the neurons in the next layer, and the final output layer outputs the final image classification based on the classifier. A common classifier is the softmax function, which normalises the feature map vectors to values between [0, 1] for each category.

### 2.2.4 Activation Function

The activation functions transfer linear functions into nonlinear operations so that the ANN can solve more complex problems, such as a nonlinear classification problem.

Common activation functions include the sigmoid, tanh, and rectified linear units (ReLU) functions, as shown in Fig. 5. Among these functions, the most popular is the ReLU function, which has been proven to have the same or better performance than the sigmoid and tanh functions [47, 48]. Moreover, it not only avoids the vanishing gradient problem but also improves the complexity of time and space with a lower computational cost.

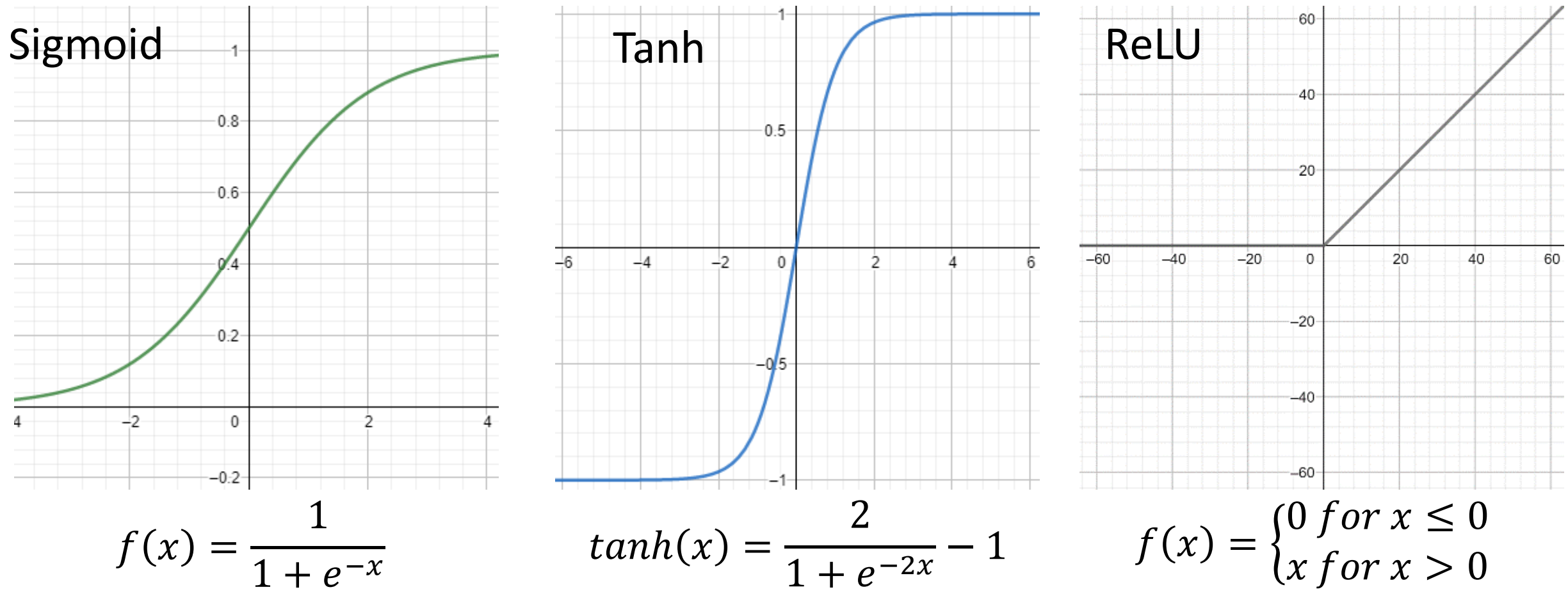

**Figure 5.** Common activation functions

### 2.3 LeNet

Since Yann LeCun proposed LeNet-1 in 1989, he continued revising it and finally proposed LeNet-5 to solve the problem of handwriting recognition in 1998; he also proposed the MNIST dataset comprising handwritten digits, which was successfully applied to the U.S. postal handwriting code recognition [49]. The LeNet-4 model is shown in Fig. 2 [50].

**Table 1.** LeNet-4 structure and hyperparameters

| Layers | Hyperparameters |
| --- | --- |
| Convolution Layer (C1) | Number of kernels: 4<br>Size of kernels: 5×5<br>Strides: 1×1 |
| Pooling Layer (P2) | Size of pooling: 2×2<br>Strides: 2×2 |
| Convolution Layer (C3) | Number of kernels: 16<br>Size of kernels: 5×5<br>Strides: 1×1 |
| Pooling Layer (P4) | Size of pooling: 5×5<br>Strides: 1×1 |
| Convolution Layer (C5) | Number of kernels: 120<br>Size of kernels: 5×5<br>Strides: 1×1 |
| Fully Connected Layer (FC6) | 120 units |
| Output Layer | 10 classifications |

First, a handwritten digital image is input, which is then convolved three times via C1, C3, and C5 and pooled twice (P2 and P4), followed by a fully connected layer (FC6). Finally, the output layer outputs the digital category from 0 to 9. The LeNet-4 structure and hyperparameters are presented in Table 1.

## 2.4 Simplified Swarm Optimisation

The SSO algorithm was proposed by Yeh in 2009 [51], and it is known to be the simplest machine learning method. SSOs have been widely applied in many fields, such as redundancy allocation problems [38], data mining [42], health care management [46], and disassembly sequencing problems [43] [52].

Let $X_i = (x_{i,1}, x_{i,2}, \ldots, x_{i,\mathrm{Nvar}})$ denote the solution $i$ and $x_{i,j}$ be its $j$th variable, where $\mathrm{N_{var}}$ is the number of variables. Assign *pBest* $P_i = (p_{i,1}, p_{i,2}, \ldots, p_{i,\mathrm{Nvar}})$, which is the personnel leader, to be the best solution among its own evolutionary history, and *gBest* $G = P_{gBest} = (G_1, G_2, \ldots, G_{\mathrm{Nvar}})$, which is the global leader, to be the best solution among all others.

SSO is a population-based stochastic optimisation technique, a swarm intelligence method, and an evolutionary computing technique. The swarm intelligence algorithm follows the leaders, that is, $G = (G_1, G_2, \ldots, G_{\mathrm{Nvar}})$ and $P_i = (p_{i,1}, p_{i,2}, \ldots, p_{i,\mathrm{Nvar}})$, to update the solutions; the most important operation in evolutionary computing is the update mechanism, which iterates continuously to obtain a solution that is close to the optimal solution.

The updating mechanism of SSO is the stepwise function listed in Eq. (2):

$$x_{i,j} = \begin{cases} g_j & \text{if } \rho_{[0,1]} \in [0, c_g = C_g) \\ p_{i,j} & \text{if } \rho_{[0,1]} \in [c_g, c_g + c_p = C_p) \\ x_{i,j} & \text{if } \rho_{[0,1]} \in [c_g + c_p, c_g + c_p + c_w = C_w) \\ x & \text{if } \rho_{[0,1]} \in [c_g + c_p + c_w, 1] \end{cases} \tag{2}$$

The update of each $x_i$ depends on one random variable $\rho_{[0,1]}$ generated uniformly between 0 and 1. $c_g = C_g \geq c_p = (C_p - C_g) \geq c_w = (C_w - C_p) \geq c_r = (1 - C_w)$ are the probabilities that the newly updated $x_{i,j}$ is equal to $g_j$, $p_{i,j}$, $x_{i,j}$ (no change) and a random feasible value $x$, respectively. Note that $c_g + c_p + c_w + c_r = 1$.

This four-term stepwise function update mechanism is efficient in balancing the exploration and exploitation abilities.

## 3. MAJOR COMPONENTS OF THE SSO-LeNet

This section introduces the major components of the proposed SSO-LeNet, including the solution structure to encode the CNN network structure, the fitness for SSO-LeNet to learn to improve itself, SDVR to adjust feasible region self-adaptively, and small-sample design matrix to tune the SSO parameters $C_g$, $C_p$, and $C_w$ systematically and efficiently. The last two components are novel to SSO-LeNet.

### 3.1 Solution Structure

The CNN structure consists of hyperparameters including the number and size of the kernels in each convolution layer, the size of the stride, and the size of the kernels in the pooling layer. Each solution in the proposed SSO-LeNet is the same as the hyperparameter settings of the CNN. Hence, the solution encoding is based on the original structure of LeNet without the need to add or delete a layer.

As shown in Figs. 2 and 6, there are 16 variables in LeNet. Thus, each solution, for example, $X = (x_1, x_2, \dots, x_{16})$, represents the 16 hyperparameters of LeNet. Their meanings and their value ranges are provided in Table 2.

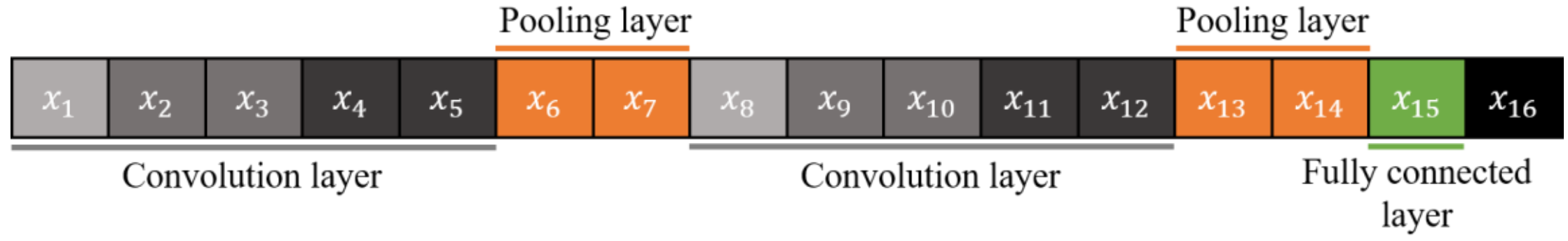

**Figure 6.** Hyperparameters encoding of the LeNet structure

**Table 2.** Meanings and range of values of hyperparameters

| Variable | Symbol | Hyperparameter | Range |
|---|---|---|---|
| $x_1$ | $N_1$ | The number of kernels of the first convolution layer. | [16, 24, 32, 40, 48, 52, 64] |
| $x_2$ | $K_{1,x}$ | The x-axis size, i.e., the number of columns of kernels of the first convolution layer. | [2 – min{11, $\text{Input}_x$}] |
| $x_3$ | $K_{1,y}$ | The size of the *y*-axis, i.e., the number of rows of kernels of the first convolution layer. | [2 – min{11, $\text{Input}_y$}] |
| $x_4$ | $S_{1,x}$ | The stride of the *x*-axis of the first convolution layer. | [1 – 4] |
| $x_5$ | $S_{1,y}$ | The stride of the *y*-axis of the first convolution layer. | [1 – 4] |
| $x_6$ | $P_{1,x}$ | The x-axis size of the first pooling layer. | [1 – $O_1^x$] |
| $x_7$ | $P_{1,y}$ | The size of the *y*-axis of the first pooling layer. | [1 – $O_1^y$] |
| $x_8$ | $N_2$ | The number of kernels of the second convolution layer. | [16, 24, 32, 40, 48, 52, 64] |
| $x_9$ | $K_{2,x}$ | The x-axis size of kernels of the second convolution layer. | [1 – $O_2^x$] |
| $x_{10}$ | $K_{2,y}$ | The size of the *y*-axis of kernels of the second convolution layer. | [1 – $O_2^y$] |
| $x_{11}$ | $S_{2,x}$ | The stride of the *x*-axis of the second convolution layer. | [1 – min(4, $O_2^x$)] |
| $x_{12}$ | $S_{2,y}$ | The stride of the y-axis of the second convolution layer. | [1 – min(4, $O_2^y$)] |
| $x_{13}$ | $P_{2,x}$ | The x-axis size of the second pooling layer. | [1 – $O_3^x$] |
| $x_{14}$ | $P_{2,y}$ | The size of the y-axis of the second pooling layer. | [1 – $O_3^y$] |
| $x_{15}$ | $U$ | The units of a fully connected layer. | [50 – 150] |
| $x_{16}$ | $b$ | The size of a training batch. | [10 – 30] |

In Table 2, $\text{Input}_x$ and $\text{Input}_y$ are the sizes of the *x*- and *y*-axes of the input images, respectively. The notation $N_i$ represents the number of kernels in the *i*th convolution layer.

The symbols *K*, *P*, and *S* denote the kernel, pooling, and stride, respectively. The first subscript, say *i*, implies that it belongs to the *i*th convolution layer or pooling layer depending on the capital letter. The second subscript, $j$, indicates that it is related to the $x$-axis or $y$-axis. For example, $K_{i,x}$ denote the x-axis sizes of the kernels in the $i$th convolution layer.

## 3.2 Fitness Function

According to Section 3.1, each solution represents a hyperparameter configuration that is a CNN structure. These hyperparameters are trained by executing LeNet and learn to improve from the fitness function, which denotes the accuracy of the testing data for each solution and is formulated as follows:

$$F_{test} = \sum_{i=1}^{N_{test}} \frac{a_i}{N_{test}} \tag{3}$$

where

$$a_i = \begin{cases} 1 & \text{if the } i\text{th sample predicted correctly} \\ 0 & \text{otherwise} \end{cases} \tag{4}$$

$N_{test}$ represents the size of testing data.

## 3.3 Sequential Dynamic Variable Range

In a traditional SSO, each solution is initialised feasibly and randomly. Owing to the operation of the CNN, the feature map becomes increasingly smaller. Hence, to ensure that the output size of the previous layer does not equal the input size of the current layer, a novel mechanism called the SDVR is proposed.

In each solution of the proposed algorithm, variables are updated one by one, as in most SSOs. However, unlike most SSOs, the feasible range of the next variable, $x_i$, is dependent on the value of the current variable, $x_{i-1}$, in the proposed algorithm.

For example, if the size of the input image is 28 × 28, and we have generated the first five variables, which are 52-8-11-1-1, then the sixth variable representing the $x$-axis size of the first pooling layer, as shown in Table 2, has to be larger than 1 and smaller than the output size of the last convolution layer $O_1^x$. Thus, we know that the sixth variable can only be randomly generated within [1, 21] based on Eq. (1).

### 3.4 Small-sample Design Matrix

Not only do the hyperparameters in LeNet need to be tuned but also the SSO parameters and variables in Eq. (2). To achieve the above two goals simultaneously, in this study, we applied three different parameter settings, as listed in Table 3.

**Table 3.** Design matrix.

| Row | $C_g$ | $C_p$ | $C_w$ |
|---|---|---|---|
| 1 | 0.4 | 0.7 | 0.9 |
| 2 | 0.5 | 0.5 | 0.8 |
| 3 | 0.5 | 0.7 | 0.7 |

In row 1, three parameters are under normal settings such that $C_g \neq C_p \neq C_w$ and $c_g = 0.4$, $c_p = 0.3$, $c_w = 0.2$, and $c_r = 0.1$, without removing any item from Eq. (2). Note that $c_g$ and $c_r$ are smaller than those in the other settings, for example, $c_g = 0.5$ and $c_r = 0.2$ in row 2. Row 2 sets $C_g = C_p = 0.5$, with a medium probability of $c_r = 0.2$. The goal of row 2 is to test whether the result is better if the role of *pBest* and a medium value of $c_r$ are removed. Row 3 assigns $C_p = C_w = 0.7$ to remove the third item from Eq. (2) with a larger value of $c_r = 0.3$, which has a higher chance of escaping local traps but a weak convergence to the optimum.

Deep learning requires a certain amount of time to train big data. In addition, SSO requires at least 30 runs and a certain number of generations and solutions to update the solutions for a better result. In the proposed SSO-LeNet, one solution update is to execute LeNet based on the structure encoded in the solution. For example, if $N_{run} = 10$, $N_{gen} = 20$, and $N_{sol} = 4$, we need to run LeNet for 800 times (10 × 20 × 4).

Hence, to reduce the training time and to find a better set of parameters, in the small sample, we only used five runs to train different SSO parameter settings and evaluated the accuracy to obtain the best configuration for this model. Specifically, to evaluate the performance of these three rows with a more robust result, a small sampling test and one-way ANOVA test were performed. Each row corresponded to an SSO with the setting shown in the design matrix. Each SSO was executed in five runs, that is, a small sampling test was conducted on three benchmark datasets: MNIST, Fashion-MNIST, and Cifar10.

The details of the small sampling test and one-way ANOVA test are provided in Section 5.2.

## 4. PROPOSED SSO-LENET

This section describes the detailed procedure of SSO-LeNet and illustrates the process through pseudocode and a flowchart.

### 4.1 Update Mechanism of the SSO-LeNet

In machine learning, parameters are the most important elements in both the update and selection procedures. The tuning of the parameters significantly affects the results. In SSOs, there are two possible methods to improve the quality of a solution: parameter-tuning and item-tuning. The former changes the value of the parameter, while the latter adds or removes items, such as Eqs. (5) and (6) removing the second and third items from Eq. (2), respectively. In this study, we used both methods to improve the quality of the solution.

$$x_{i,j} = \begin{cases} g_j & \text{if } \rho_{[0,1]} \in [0, c_g = C_g) \\ x_{i,j} & \text{if } \rho_{[0,1]} \in [c_g + c_p, c_g + c_p + c_w = C_w) \\ x & \text{if } \rho_{[0,1]} \in [c_g + c_p + c_w, 1] \end{cases} \tag{5}$$

$$x_{i,j} = \begin{cases} g_j & \text{if } \rho_{[0,1]} \in [0, c_g = C_g) \\ p_{i,j} & \text{if } \rho_{[0,1]} \in [c_g, c_g + c_p = C_p) \\ x & \text{if } \rho_{[0,1]} \in [c_g + c_p + c_w, 1] \end{cases} . \tag{6}$$

### 4.2 Stopping Criteria of SSO-LeNet

All algorithms have different stopping criteria according to different conditions. As mentioned in Section 3.4, a CNN always takes a long time to execute. To stop the execution of the algorithm earlier, is the generation in which the accuracy obtained from the proposed algorithm starts better than that of LeNet is considered as the stopping criterion.

### 4.3 Pseudocode and Flowchart

Let $N_{gen}$, $N_{sol}$, and $N_{var}$ be the number of generations, solutions, and variables, respectively. We assume that Z is the solution obtained from LeNet. The pseudocode of the proposed SSO-LeNet is as follows.

**STEP 0.** Generate a solution of hype-parameters $Z$ and calculate its fitness $F(Z)$ using LeNet.

**STEP 1.** Initialise solutions $P_i = X_i$, calculate $F(P_i) = F(X_i)$ for $i = 1, 2, \ldots, N_{sol}$, find $G = F(P_{gBest})$, and let $t = i = 1$.

**STEP 2.** Update $X_i$ based on the best parameter setting obtained using the small sampling discussed in Section 3.4 and calculate $F(X_i)$.

**STEP 3.** If $F(X_i)$ is better than $F(P_i)$, then $P_i = X_i$. Otherwise, return to STEP 5.

**STEP 4.** If $F(X_i)$ is better than $F(G)$, let $G = X_i$.

**STEP 5.** If $i < N_{sol}$, then $i = i + 1$ and return to STEP 2.

**STEP 6a.** If $F(G)$ is better than $F(Z)$, then stop.

**STEP 6.** If $t < N_{gen}$, then $t = t + 1$, $i = 1$, and return to STEP 2. Otherwise, stop.

Note that STEP 6a is an optional step and an early stopping criterion. When we implement STEP 6a, the algorithm ends earlier. A flowchart of the proposed algorithm is shown in Fig. 7.

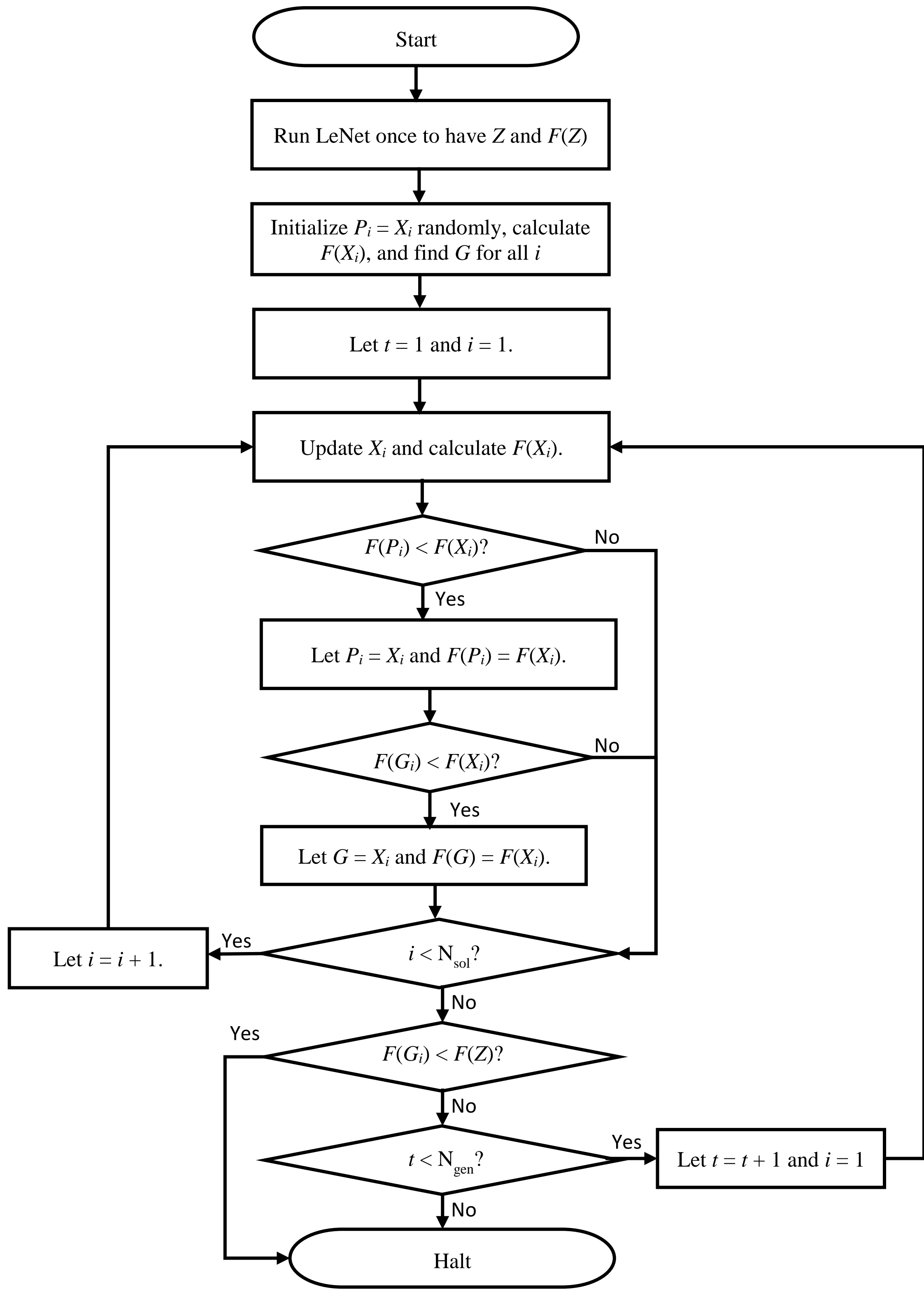

**Figure 7.** The flowchart of the proposed algorithm.

## 5. EXPERIMENTS

Three experiments, Ex1, Ex2, and Ex3, were conducted on three benchmark datasets: MNIST, Fashion-MNIST, and Cifar10. Ex1 mainly tuned the parameters of the SSO using the small-sample design matrix listed in Table 4. Based on the best SSO parameters selected in Ex1, Ex2 and Ex3 were conducted to demonstrate the performance of the proposed SSO-LeNet by comparing it to the traditional LeNet [49] and PSO-LeNet [39], respectively.

### 5.1 Three Datasets and Experiment Environments

Three datasets, namely MNIST, Fashion-MNIST, and Cifar10 [49], were used in the experiments, and they are summarised as follows.

1. MNIST is a dataset of handwritten digits, 0–9, with 10 categories, as shown in Fig. 8(1). It collects 70,000 handwriting images, comprising 60,000 training images and 10,000 testing images. Each image is 28 × 28 pixels with different shades of grayscales.
2. The Fashion-MNIST dataset is an advanced MNIST dataset. It also contains 70,000 images with 28 × 28 pixels, of which 60,000 images are for training and 10,000 images are for testing. However, the categories included in Fashion-MNIST are T-shirts, trousers, coats, and bags, as shown in Fig. 8(2).
3. The CIFAR-10 dataset comprises animals and vehicles with 60,000 32 × 32 colour images in 10 classes. As shown in Fig. 8(3), there are 50,000 training images and 10,000 test images.

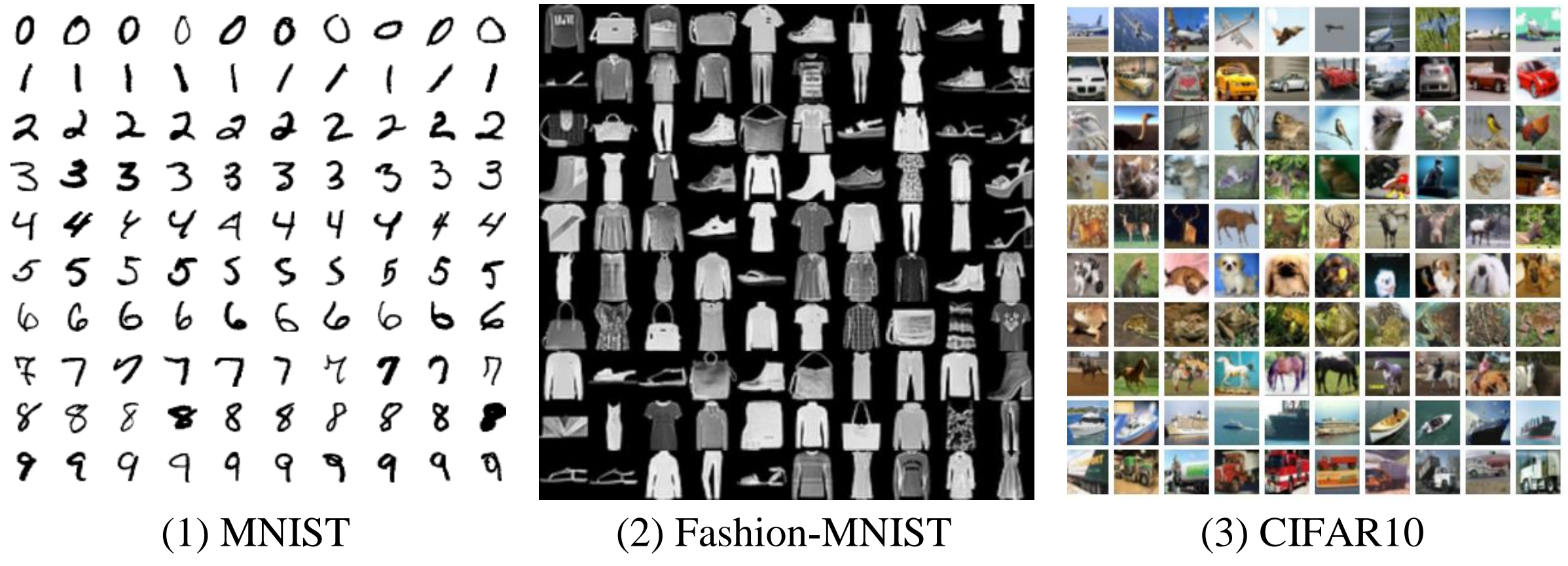

(1) MNIST (2) Fashion-MNIST (3) CIFAR10

**Figure 8.** Three datasets.

The proposed SSO-LeNet, original LeNet [49], and PSO-LeNet [39] were all coded using Python3.7.9 and Tensorflow2.1 in the Spyder and run on Intel Core i9-9900K CPU @ 3.6 GHz, 48 GB of memory, and an NVIDIA GeForce RTX 2070 GPU on the above three datasets.

## 5.2 Ex1: Tune SSO Parameters and Items

In Ex1, the SSO parameters were tuned based on the design matrix listed in Table 4 using the small-sample concept conducted on the three datasets [39, 49].

Table 4 shows the best and mean accuracy obtained by implementing SSO-LeNet for five runs under 20 generations and 50 solutions, that is, $N_{run}$ = 5, $N_{gen}$ = 20, and $N_{sol}$ = 50, for three different rows of $C_g$, $C_p$, and $C_w$, respectively, as shown in Table 3.

Let the parameters, $C_g$, $C_p$, and $C_w$, of the *i*th row in the designed matrix be SSO*i* and $N_{best}$ and $T_{best}$ be the number of generations and the runtime that starts to have a better accuracy than that of the original LeNet, respectively. Figs. 9(1)–9(3) show the boxplots that demonstrate the accuracy, $N_{best}$, and $T_{best}$ under three different parameter settings.

**Table 4.** The best and mean of accuracy for different configurations.

| Parameter configurations | Best accuracy | Mean accuracy |
|---|---|---|
| SSO1 | **0.9925000072** | **0.9905920005** |
| SSO2 | 0.9922000170 | 0.9902770066 |
| SSO3 | 0.9923999906 | 0.9903220028 |

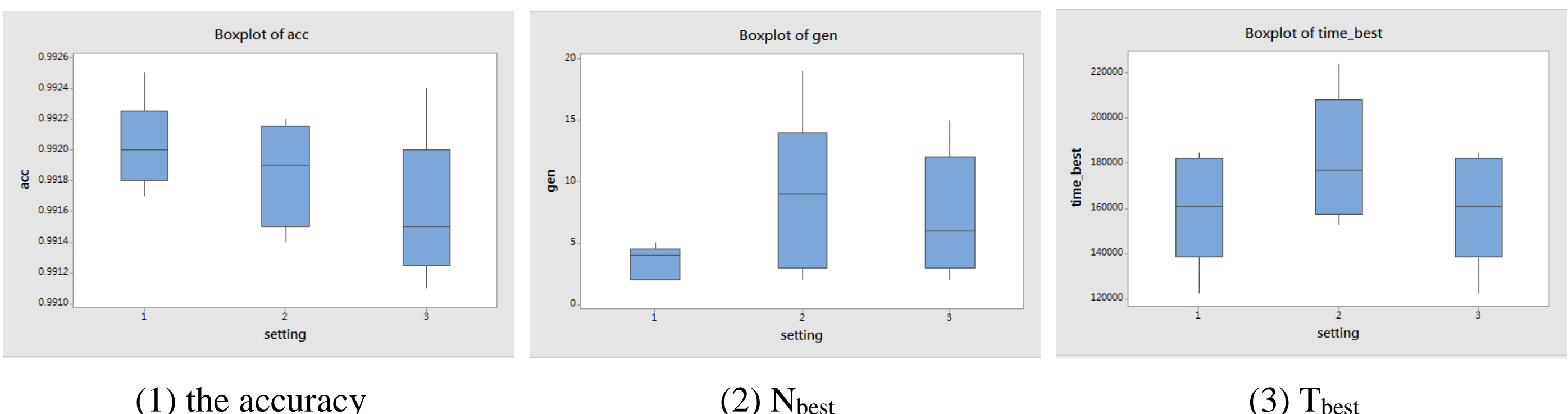

(1) the accuracy (2) $N_{best}$ (3) $T_{best}$

**Figure 9.** The boxplots of under three settings.

From Table 4 and Fig. 9, the first row, that is, ($C_g$, $C_p$, $C_w$) = (0.4, 0.7, 0.9), has the best accuracy, mean accuracy, $N_{best}$, and $T_{best}$. In addition, the ranges of accuracy, $N_{best}$, and $T_{best}$ are more stable than the others, as shown in Fig. 9.

From the normality, homoscedasticity, and independence tests, as depicted in Fig. 10, the experimental results meet the three assumptions of ANOVA. Hence, a one-way ANOVA test can be implemented to check whether there is any significant difference among the three designs for the SSO parameter settings.

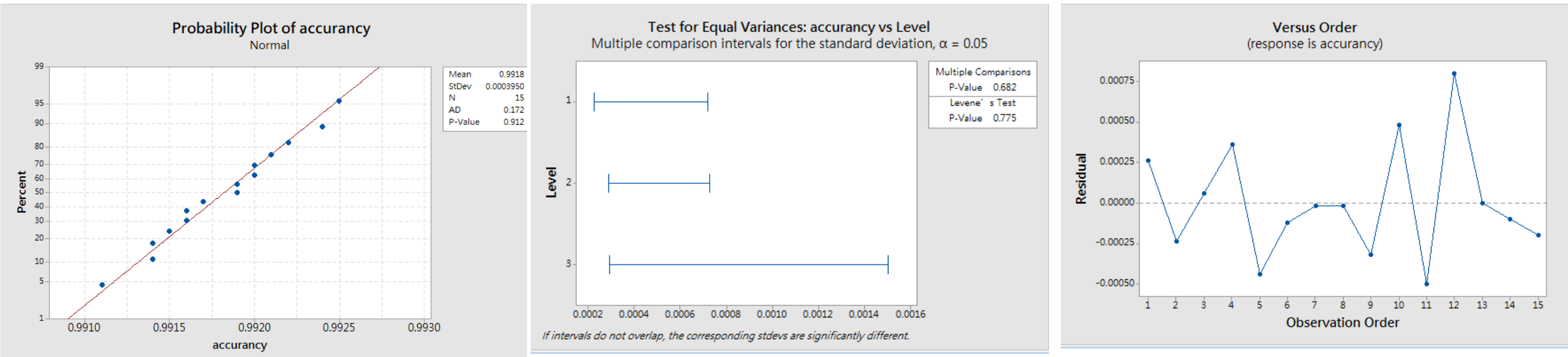

(1) The normality test (2) the homoscedasticity test (3) the independence test

**Figure. 10** The normality, homoscedasticity, and independence tests.

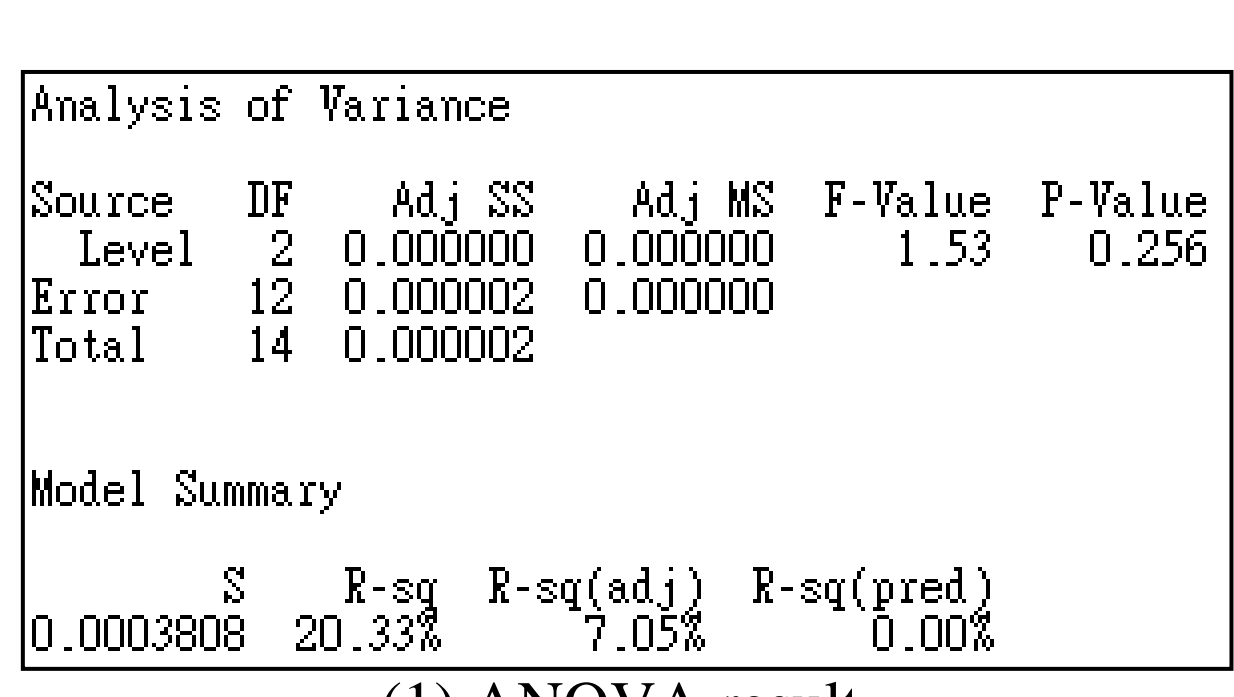

Analysis of Variance

| Source | DF | Adj SS | Adj MS | F-Value | P-Value |
|---|---|---|---|---|---|
| Level | 2 | 0.000000 | 0.000000 | 1.53 | 0.256 |
| Error | 12 | 0.000002 | 0.000000 | | |
| Total | 14 | 0.000002 | | | |

Model Summary

| S | R-sq | R-sq(adj) | R-sq(pred) |
|---|---|---|---|
| 0.0003808 | 20.33% | 7.05% | 0.00% |

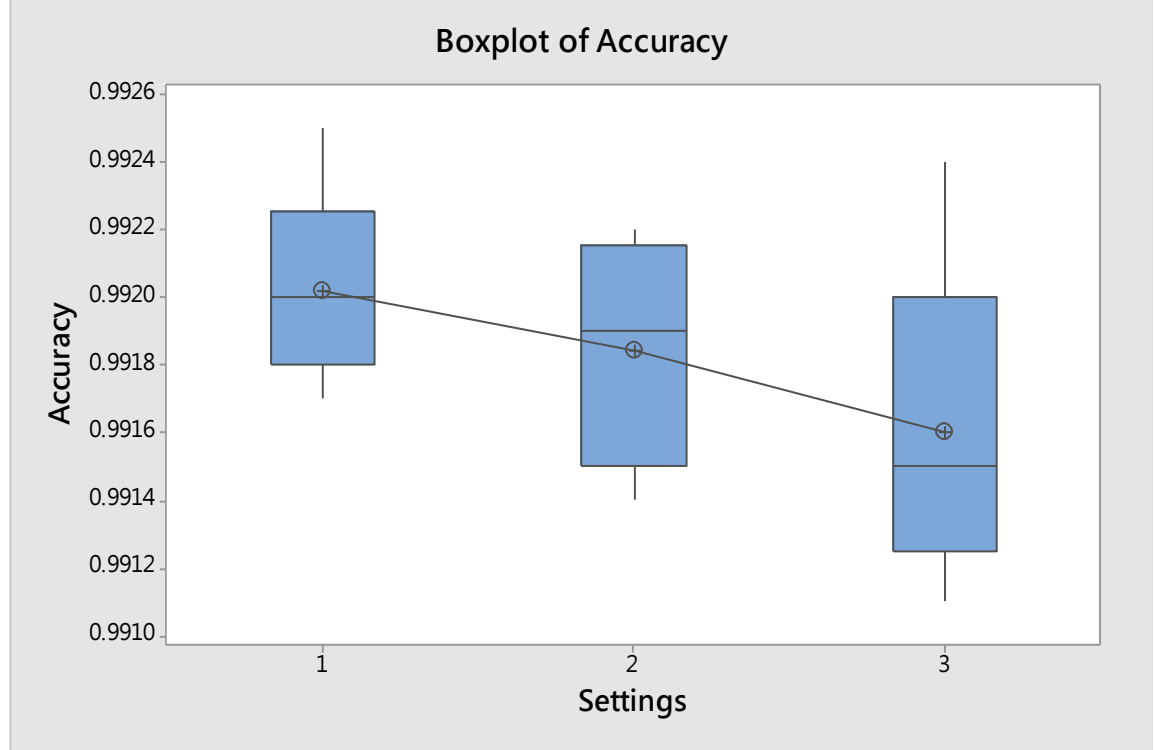

(1) ANOVA result. (2) The accuracy boxplot under 3 settings.

**Figure 11.** ANOVA test.

The ANOVA results are shown in Fig. 11(1), and the boxplots under different rows are shown in Fig. 11(2). The ANOVA table (Fig. 11(1)) shows that the p-value is 0.256 > 0.05, where the confidence level is 95%, that is, the variances of the rows are not significantly unequal. Hence, the first design ($C_g$, $C_p$, $C_w$) = (0.4, 0.7, 0.9) was selected for Ex2 and Ex3 because it had the best accuracy, mean accuracy, $N_{best}$, and $T_{best}$.

### 5.3 Ex2: Compare to LeNet

Ex2 was conducted to validate whether there is an improvement in LeNet after using the proposed SSO-LeNet. According to the results in Section 5.1, the best parameters ($C_g$, $C_p$, $C_w$) = (0.4, 0.7, 0.9) were adopted in Ex2, and the other settings are summarised in Table 5.

**Table 5.** Summary of parameters

| Parameter | Value |
|---|---|
| ($C_g$, $C_p$, $C_w$) | (0.4, 0.7, 0.9) |
| $N_{run}$ | 30 |
| $N_{gen}$ | 20 or the generation of which accuracy is larger than that of LeNet |
| $N_{sol}$ | 30 |
| Epoch | 10 |
| Activation function | ReLU |
| Classifier | Softmax |
| Optimizer | the Stochastic Gradient Descent (SGD) |
| Loss function | Cross entropy |

The maximal accuracy $F_{max}$, minimum accuracy $F_{min}$, mean accuracy $F_{mean}$, standard deviation of accuracy $F_{std}$, mean testing runtime $T_{test}$, and total training runtime $T_{train}$ obtained from the original LeNet and the proposed SSO-LeNet for the three datasets are listed in Table 6.

**Table 6.** The results of the dataset.

| Dataset | Method | $F_{max}$ | $F_{min}$ | $F_{mean}$ | $F_{std}$ | $T_{test}$ | $T_{train}$ |
|---|---|---|---|---|---|---|---|
| MNIST | LeNet | 0.9918 | 0.9876 | 0.9899 | 0.0009 | 203982.81 | **6119484.375** |
| | SSO-LeNet | **0.9923** | **0.9891** | **0.9905** | **0.0008** | **172894.79** | 916772812.5 |
| Fashion-MNIST | LeNet | 0.9005 | 0.8917 | 0.9001 | **0.0036** | 219691.15 | **6590734.38** |
| | SSO-LeNet | **0.9113** | **0.8933** | **0.9021** | 0.0039 | **192489.50** | 1229022948 |
| Cifar10 | LeNet | 0.6925 | 0.6596 | 0.6769 | 0.0092 | 2181456.25 | **65443687.5** |
| | SSO-LeNet | **0.6951** | **0.6601** | **0.6807** | **0.0079** | **1899773.44** | 16671130781 |

From Table 6, it can be observed that all values obtained from SSO-LeNet are better than those from LeNet except $T_{train}$ for all datasets and $F_{std}$ for MNIST. However, the difference between $\mathbf{F_{std}}$ and MNIST was less than 0.003. In addition, it gives full play to the idea that fiscal reserves should be used when the occasion calls for it, that is, the time taken for testing is more important than that for training.

Hence, the proposed SSO-LeNet improves these two values in terms of accuracy and runtime.

An interesting phenomenon is that the kernel in the best solution is not always a square matrix. In general, each kernel (filter) is a square matrix in a CNN. However, Table 7 shows that all input images have a square shape initially, that is, all runs have a square shape for the three datasets. After performing more convolution and pooling layers for each generation in the proposed SSO-LeNet, the size of the output feature map is close to a rectangle such that its number of columns (i.e., the *y*-axis size of kernels) is less than that of rows (i.e., the *x*-axis size of kernels), for example, 18 out of 30 runs with $y < x$ for the MNIST. The above phenomenon can be used to improve the convergence rate of the proposed SSO-LeNet in the future.

**Table 7.** Output size of the feature maps.

| **Dataset** | **Size*** | **Input Image** | **$C_1^{\#}$ output** | **$P_1^{\&}$ output** | **$C_2$ output** | **$P_2$ output** |
|---|---|---|---|---|---|---|
| MNIST | $x>y$ | 0 | 11 | **20** | **18** | **18** |
| | $x<y$ | 0 | **16** | 10 | 12 | 11 |
| | $x=y$ | **30** | 3 | 0 | 0 | 1 |
| Fashion-MNIST | $x>y$ | 0 | **16** | **24** | **23** | **22** |
| | $x<y$ | 0 | 7 | 3 | 3 | 4 |
| | $x=y$ | **30** | 7 | 3 | 4 | 4 |
| Cifar10 | $x>y$ | 0 | **11** | **24** | **25** | **24** |
| | $x<y$ | 0 | **11** | 3 | 3 | 5 |
| | $x=y$ | **30** | 8 | 3 | 2 | 1 |

[*]: The *x*-axis and *y*-axis sizes of kernels denote by *x* and y, respectively.
[#]: $C_i$ is the *i*th convolution layer.
[#]: $P_i$ is the *i*th pooling layer.

## 5.4 Ex3: Compare with PSO-LeNet

There is another existing algorithm called PSO-LeNet [39] based on the machine learning PSO to improve LeNet [49]. To further demonstrate the performance of the proposed SSO-LeNet, the proposed SSO-LeNet is compared to PSO-LeNet [39].

For a fair comparison, the setting of SSO-LeNet is based on that used in PSO-LeNet [39], as shown in Table 8.

**Table 8.** Parameters of the two algorithms.

| Parameters | Value |
|---|---|
| ($C_g$, $C_p$, $C_w$) | (0.4, 0.7, 0.9) |
| ($W$, $C_1$, $C_2$) | (0.5, 0.5, 0.5) |
| $N_{run}$ | 30 |
| $N_{gen}$ | 100 |
| $N_{sol}$ | 5 |
| Batch size | 128 |
| Epoch | 100 |
| Activation function | ReLU |
| Classifier | Softmax |
| Optimizer | Adam |
| Loss function | Cross entropy |

The accuracy and runtime of the best solution with the highest accuracy are listed in Table 9 for the three datasets.

**Table 9.** Comparisons between SSO-LeNet and PSO-LeNet.

| **Dataset** | **SSO-LeNet** | | **PSO-LeNet** | |
|---|---|---|---|---|
| | **Accuracy** | **Time(ms)** | **Accuracy** | **Time(ms)** |
| MNIST | **0.9958** | 1,016,953.13 | 0.9940 | **569,015.63** |
| Fashion-MNIST | **0.9275** | 1,298,375.00 | 0.9267 | **984,109.38** |
| Cifar10 | **0.7313** | **397,187.50** | 0.7016 | 1,203,890.63 |

As can be observed, the accuracy of the proposed SSO-LeNet is better than that of PSO-LeNet [39] for all datasets. In contrast, PSO-LeNet has a better runtime in MNIST and Fashion-MNIST.

Different hyperparameter configurations may require different times to obtain the best results. However, the accuracy of detecting defective products is important to manufacturers, and even a difference of 0.1% in accuracy can have a significant impact on mass production. Moreover, once the model is trained, the accuracy can be higher, and production efficiency can be improved.

Consequently, after training, the proposed SSO-LeNet outperformed PSO-LeNet [39].

## 6. CONCLUSIONS

CNNs have drawn research attention because image recognition is one of the most important applications in people’s lives. Hyperparameter optimisation is an economical and convenient way to improve the accuracy of CNNs. A new algorithm called SSO-LeNet was proposed to optimise the hyperparameters of the LeNet-4 model in a CNN.

The parameters of the proposed SSO-LeNet were tuned using a design matrix on small samples. A comparison with the traditional LeNet [49] on three benchmark datasets revealed that SSO-LeNet outperforms the original LeNet in terms of both accuracy and testing time. In addition, the proposed SSO-LeNet outperforms PSO-LeNet [39] in terms of accuracy.

In the future, this algorithm can be applied to other CNN models, such as AlexNet, VGG, and GoogLeNet. It provides a new method for optimising the hyperparameters to obtain better results with the existing model architecture.

## ACKNOWLEDGEMENT

This research was supported in part by the Ministry of Science and Technology of Taiwan (MOST 107-2221-E-007-072-MY3). This article was once submitted to arXiv as a temporary submission that was just for reference and did not provide the copyright.